\documentclass[default]{sn-jnl}% Default
\usepackage{xcolor}
\usepackage{bbm}
\usepackage{subcaption}
\usepackage{rotating}
\usepackage{multirow}
\usepackage{array}
\newcolumntype{C}[1]{>{\centering\arraybackslash}p{#1}} % centered version of 'p' column type
\usepackage{geometry}
\usepackage{pdflscape}
\usepackage{lipsum}
\usepackage{amsmath,amsfonts,amssymb,stmaryrd}
\usepackage{amssymb}
\usepackage{pifont}
\usepackage{amssymb}
\usepackage{subcaption}
\usepackage{amsmath}
\usepackage{accents}
\usepackage{gensymb}
\usepackage{mathtools}
\usepackage{xcolor}
\usepackage{longtable}
\usepackage{verbatim}
\usepackage{amsthm}
\usepackage{acronym}
\usepackage{xr}
\usepackage[utf8]{inputenc}
\usepackage{algpseudocode}
\usepackage{eucal}
\usepackage[algo2e]{algorithm2e} 
\usepackage{float}
\usepackage{xr}
\DeclarePairedDelimiter{\abs}{\lvert}{\rvert}

\makeatletter
\newcommand*{\addFileDependency}[1]{
  \typeout{(#1)}
  \@addtofilelist{#1}
  \IfFileExists{#1}{}{\typeout{No file #1.}}
}

\makeatother

\newcommand*{\myexternaldocument}[1]{
    \externaldocument{#1}
    \addFileDependency{#1.tex}
    \addFileDependency{#1.aux}
}

\newcommand{\proto}{MAFUS}

\myexternaldocument{appendix}

\acrodef{DL}[DL]{Deep Learning}
\acrodef{AI}{Artificial intelligence}
\acrodef{MAFLD}{Metabolic (dysfunction) associated fatty liver disease}
\acrodef{NAFLD}{Nonalcoholic Fatty Liver Disease}
\acrodef{OS}{Overall Survival}
\acrodef{DBP}{Diastolic Blood Pressure}
\acrodef{TC}{Total Cholesterol}
\acrodef{HDL-C}{High-Density Lipoprotein Cholesterol}
\acrodef{LDL-C}{Low-Density Lipoprotein Cholesterol}
\acrodef{GGT}{Glutamyltransferase}
\acrodef{AST}{Aspartate Aminotransferase}
\acrodef{CI}{Confidence Interval}
\acrodef{BMI}{Body Mass Index}
\acrodef{MLP}{Multilayer Perceptron}
\acrodef{RF}{Random Forest}
\acrodef{SVM}{Support Vector Machine}
\acrodef{DT}{Decision Tree}
\acrodef{DNN}{Deep Neural Network}
\acrodef{ML}{Machine Learning}
\acrodef{MD}{Metabolic Dysregulation}
\acrodef{CVD}{Cardiovascular Diseases}
\acrodef{DSD}{Digestive System Disease-related mortality}
\acrodef{IQR}{Interquartile Range}
\acrodef{SD}{Standard Deviation}
\acrodef{CI}{Confident Interval}
\acrodef{HR}{Hazard Ratio}
\acrodef{MSE}{Mean Square Error}
\acrodef{PH3}{Parametric Hazard with 3 degrees of freedom}
\acrodef{CKD}{Chronic Kidney Disease}
\acrodef{ROC}{Receiver Operating Characteristics}
\acrodef{HOMA}{HOMA-IR Index}
\acrodef{PBI}[PBI]{Patient Behaviour Index}
\acrodef{TAI}[TAI]{Therapy Adherence Index}
\acrodef{HBI}[HBI]{Healthy Behaviour Index}
\acrodef{CP}[CP]{Clinical Pathway}
\acrodef{IoMT}[IoMT]{Internet of Medical Things}
\acrodef{IoT}[IoT]{Internet of Things}
\acrodef{EHR}[EHR]{Electronic Health Records}
\acrodef{CG}[CG]{Clinical Guidelines}
\acrodef{BMK}[BMK]{Basic Medical Knowledge}
\acrodef{BPMN}[BPMN]{Business Process Modelling Notation}
\acrodef{ACS}[ACS]{Acute Coronary Syndrome}
\acrodef{aHR}[aHR]{adjusted Hazard Ratio}
\acrodef{FIB-4}[FIB-4]{Fibrosis-4}
\acrodef{MSE}[MSE]{Mean Square Error}
\acrodef{MAE}[MAE]{Mean Absolute Error}
\acrodef{ReLU}[ReLU]{Rectified Linear Unit}
\acrodef{SBP}[SBP]{Systolic blood pressure}
\acrodef{GOT}[AST]{Aspartate aminotransferase}
\acrodef{GPT}[ALT]{Alanine aminotransferase}
\acrodef{DMLC}[DMLC]{Distributed Machine Learning Community}
\acrodef{GOSS}[GOSS]{Gradient-based One-Side Sampling}
\acrodef{LIME}[LIME]{Local Interpretable Model-Agnostic Explanations}
\acrodef{SHAP}[SHAP]{Shapley Additive exPlanations }
\acrodef{CM}[CM]{Confusion Matrix}
\acrodef{XAI}[XAI]{Explainable Artificial Intelligence}
\acrodef{XGBoost}[XGBoost]{Extreme Gradient Boosting}
\acrodef{CM}{Confusion Matrix}

\jyear{2022}%

\theoremstyle{thmstyleone}%
\theoremstyle{thmstyletwo}%

\theoremstyle{thmstylethree}%

\begin{document}

\title[]{\proto: a Framework to predict mortality risk in MAFLD subjects}

%\title[ Deep Neural Network Approach to predict mortality risk in MAFLD subjects]{A Deep Neural Network Approach to predict mortality risk in MAFLD subjects}

%%%%%%%%%%INSERIRE ANCHE ORCID%%%%%%%%%%%%%%%%%%%%%%%%%%%%%%%%%%%%%%%%
\author*[1]{\fnm{Domenico} \sur{Lofù}}\email{domenico.lofu@poliba.it}
\author*[1]{\fnm{Paolo} \sur{Sorino}}\email{paolo.sorino@poliba.it}
\author[1]{\fnm{Tommaso} \sur{Colafiglio}}\email{tommaso.colafiglio@poliba.it}
\author[2]{\fnm{Caterina} \sur{Bonfiglio}}\email{catia.bonfiglio@irccsdebellis.it}
\author[1]{\fnm{Fedelucio} \sur{Narducci}}\email{fedelucio.narducci@poliba.it}
\author[1]{\fnm{Tommaso} \sur{Di Noia}}\email{tommaso.dinoia@poliba.it}
\author[1]{\fnm{Eugenio} \sur{Di Sciascio}}\email{eugenio.disciascio@poliba.it}
%\author[2]{\fnm{Rodolfo} \sur{Sardone}}\email{rodolfo.sardone@irccsdebellis.it}


%\affil[2]{\orgdiv{Department}, \orgname{Organization}, \orgaddress{\street{Street}, \city{City}, \postcode{10587}, \state{State}, \country{Country}}}

\affil[1]{\orgdiv{Dept. of Electrical and Information Engineering (DEI)}, \orgname{Politecnico di Bari},\\ \orgaddress{\state{Bari (BA), 70126}, \country{Italy}}}

%\affil[2]{\orgdiv{Data Sciences and Innovation National Institute of Gastroenterology}, \orgname{“S. De Bellis” Research Hospital}, \orgaddress{\state{Castellana Grotte (BA), 70013}, \country{Italy}}}

\affil[2]{\orgdiv{Laboratory of Epidemiology and Biostatistics, National Institute of Gastroenterology}, \orgname{“S de Bellis” Research Hospital}, \orgaddress{\state{Castellana Grotte (BA), 70013}, \country{Italy}}}

\abstract{
\ac{MAFLD} establishes new criteria for diagnosing fatty liver disease independent of alcohol consumption and concurrent viral hepatitis infection. However, the long-term outcome of MAFLD subjects is sparse. Few articles are focused on mortality in MAFLD subjects, and none investigate how to predict a fatal outcome.
%Our goal was to develop a machine learning algorithm based on simple anthropometric, biochemical, and other readily available variables to build an artificial intelligence-based predictive algorithm to identify mortality in MAFLD subjects.
%Our goal is to develop an artificial intelligence-based framework. To identify mortality in MAFLD subjects, we developed a machine learning algorithm based on anthropometric, biochemical, and other readily available variables.f
In this paper, we propose an artificial intelligence-based framework named MAFUS that physicians can use for predicting mortality in MAFLD subjects. The framework uses data from various anthropometric and biochemical sources based on \ac{ML} algorithms.
%Starting from the initial $25$ features in the dataset, $10$ features are selected using an XGB feature importance analysis. The feature selected are used in \ac{ML} algorithms. 
The framework has been tested on a state-of-the-art dataset on which five ML algorithms are trained. %The percentage accuracy, precision, Recall, F1 score, \ac{ROC}, and Confusion Matrix are evaluated for each algorithm to determine the best model.
%An \ac{SVM} (best model) is then developed. 
Support Vector Machines resulted in being the best model.
Furthermore, an \ac{XAI} analysis has been performed to understand the SVM's diagnostic reasoning and the contribution of each feature to the prediction.
%Then a new framework, "MAFUS", is proposed. 
%As a result of experimental settings,  SVM algorithm performed better than all other models with $82\%$ accuracy and F1 score $0.61$ and the lowest number of misclassified in class \textit{Mortality (Yes)}. An SVM approach support mortality in individuals with MAFLD. 
The MAFUS framework is easy to apply, and the required parameters are readily available in the dataset.
}

%\keywords{MAFLD, Mortality Risk, Epidemiology, Machine Learning, Deep Learning.}


\maketitle
%--------------------------------------------------------------------------------------------------------------------------------------------------------
\section{Introduction}
\label{sec:introduction}

\ac{NAFLD} is the leading cause of chronic liver disease in Western countries, as well as a condition raising the risk for cardiovascular diseases, type $2$ diabetes mellitus and chronic renal disease, and increased mortality ~\cite{fazel2016epidemiology, levene2012epidemiology}.
The prevalence of NAFLD is estimated at $24\%$ worldwide (from $15\%$ in $2005$ to $25\%$ in $2010$).
A meta-analysis published in $2016$ reported an average prevalence of $23.71\%$ in Europe~\cite{younossi2016global}. Population-based studies conducted in our geographical area (district of Bari, Apulian Region, Italy), are estimated a NAFLD prevalence of around $30\%$, mainly among male subjects and older people~\cite{cozzolongo2009epidemiology}.
NAFLD is defined as an accumulation of high triglycerides in the hepatocytes ($>5\%$ of the liver volume), after excluding viral infectious causes or other specific liver diseases~\cite{neuschwander2003nonalcoholic}.
NAFLD can manifest as fatty liver disease (hepato-steatosis) or as non-alcoholic steatohepatitis (NASH).
An evolution of the former in which steatosis comes with inflammation, hepatocellular injury, and fibrogenic activation may lead to cirrhosis and hepatocarcinogenesis. 
\ac{MAFLD} is a novel concept proposed in $2020$ aiming to replace the term \ac{NAFLD}~\cite{eslam2020mafld}.
Unlike NAFLD, MAFLD does not require the exclusion of other etiologies of liver disease, such as excessive alcohol consumption or viral hepatitis~\cite{eslam2020asian}. MAFLD is diagnosed in subjects who have both hepatic steatosis and one of the following three metabolic diseases: overweight/obesity (Subtype 1), evidence of \ac{MD} in lean subjects (Subtype 2), or diabetes mellitus (Subtype 3)~\cite{eslam2020new}. 

In addition, it is unclear whether the new definition provides better detectability regarding hard clinical endpoints (e.g., mortality). Although ample evidence of the association of MAFLD with \ac{CVD}, malignancies, and liver-related endpoints, the impact on mortality remains controversial~\cite{semmler2021metabolic}.

In recent years, due to the increasing prevalence of NAFLD and a new definition of MAFLD, there is a research trend toward identifying low-cost diagnostic methods, and \ac{ML} is acknowledged as a valuable method. In clinical practice, numerous works have shown how ML or e-health tools are considered various alternatives to standard diagnostic methods \cite{casalino2021microrna,9945542,lella2021ensemble,casalino2022evaluating} such as Magnetic Resonance Imaging (MRI), ultrasounds, etc. 
%For instance, a study conducted by Casalino \emph{et al.}~\cite{casalino2021microrna}, Castellana \emph{et al.}~\cite{9945542} and another one by Lella \emph{et al.}~\cite{lella2021ensemble} shows how AI is applied in the e-health domain. Comparatively, an additional study by Casalino \emph{et al.}~\cite{casalino2022evaluating} shows how a contactless mHealth solution is applied. 
 ML approaches are already used for NAFLD diagnosis~\cite{sorino2020selecting,sorino2021development}.

According to Curci \emph{et al.}~\cite{curci2022effect} who demonstrated the benefits of Physical Activity and Diet intervention in subjects with MAFLD, in this paper, we propose a method to predict the patient's outcome \textit{(Mortality (Yes/No))}, in order to allow the physician, to suggest a lifestyle modification (Diet and Physical Activity) to the subjects when \textit{Mortality (Yes)} is predicted.

%In this paper, we propose a method for identifying the patient's status prediction \textit{(Mortality (Yes/No))} in a way to allow early intervention when \textit{Mortality (Yes)} is predicted in MAFLD subjects. 
The experimental setup is designed on a population-based cohort after $15$ years of follow-up. A dataset containing $25$ variables is used in the experiments. A feature relevance analysis is performed to identify the features correlated to the target variable.

%In addition we compare a variety of ML algorithms based on readily available laboratory variables identified \textcolor{red}{comparing the accuracy, precision, recall, F1 score,  and confusion matrix} to find an optimal method for classifying mortality in MAFLD subjects \textcolor{red}{to develop a system that provides an accurate mortality classification based on readily available biochemical and anthropometric parameters in a cohort of MAFLD subjects, explaining to the clinician the reasoning that led the system toward a given prediction.} %\textcolor{blue}{In the next step, we apply an \ac{XAI} analysis to fully understand the algorithmic reasoning and the contribution of each variable to the prediction.}

To determine an optimal method for predicting mortality in MAFLD subjects, we compare ML algorithms based on readily available laboratory variables for accuracy, precision, recall, F1 score, and confusion matrix.
As a result, we developed a system for accurately predicting mortality based on biochemical and anthropometric parameters in a cohort of MAFLD subjects so that the clinician understood the rationale for the system's predictions (Mortality (Yes/No).

%\textcolor{blu}{(CON CIO' CHE HO SCRITTO SOPRA HO SINTETIZZATO TUTTA QUESTA PARTE SPOSTANDO L'AIM DA IN THIS PAPER WE PROPOSE FINO ALLA PARTE IN ROSSO. VALUTA UN PO MIMMO)
%The goal of this article is to train a classifier on biochemical and anthropometric parameters (Weight, Blood Glucose,\ac{HDL-C}, \ac{LDL-C}, Age, \ac{HOMA},\ac{BMI}, Gender, Total Cholesterol, Triglycerides) readily available in healthcare databases, to predict mortality in a cohort of MAFLD subjects.}

%As a result, the physicians can preemptively intervene on those parameters to change the patient's status from the \textit{Mortality (Yes)} to \textit{Mortality (No)}.
Explainability analysis is used to highlight the characteristics \textit{Mortality (Yes)} versus \textit{Mortality (No)} with relative final prediction to support the physician's decision.

The remainder of the paper is organized as follows: in Section~\ref{sec:back_and_related} we overview the most relevant related work;\\
Section~\ref{sec:proposed_approach}  describes the proposed approach and
%, including  the machine learning algorithm used, %metrics used to measure the performance of classifiers, 
%the information about the population study, the clinical data collection, and the dataset used for the experiments, (ii) explainability techniques for understanding the reasoning of algorithms, and (iii) the representation of the \proto~framework and pseudocode~\ref{alg:CMgeneration}.
Section~\ref{sec:exp_res} reports the metrics used to measure the performance of our model and the explanation obtained from the XAI analysis. 
FInally, Section~\ref{sec:conclusion} tightens the conclusion and future research directions.

%In particular it illustrates: 
%(i) information about the population study and how data were collected (Section~\ref{sec:pop}), 
%(ii) dataset used for the experiments (Section~\ref{sec:dataset}),
%(iii) preprocessing and feature relevance operations adopted (Section~\ref{sec:preproc}),
%(iv) description of the classification models used in the comparison algorithm %(Section~\ref{sec:algo_compa}), 
%(v) evaluation metrics used to measure the performance of classifiers (Section~\ref{sec:metrics}),
%(vi) explainability techniques for understanding the reasoning of algorithms %(Section~\ref{sec:explainability} and,
%(vii) framework representation with presudocode.

%\textcolor{blue}{Section~\ref{sec:stat_and_ml_analysis} provides the details of the statistical and artificial intelligence approach used. 
%Section~\ref{sec:results} reports (i) the performance assessment of the evaluation of the developed prognostic index and the performance of the statistical model, (ii) the performance of all machine learning and deep learning algorithms and (iii) the performance of the best algorithm identified.
%Section~\ref{sec:discussion} reports the performance evaluation obtained from the statistical model and the best algorithm identified, while Section~\ref{sec:conclusion} tightens conclusion and future research directions.}
%--------------------------------------------------------------------------------------------------------------------------------------------------------
\section{Related work}
\label{sec:back_and_related}

In this section, we reviewed the state of the art on the MAFLD condition and clarified the significant differences between this study and existing work by highlighting the innovative contribution of this study.
Dan-Qin Sun \emph{et al.}~\cite{sun2021mafld} investigated the distribution of MAFLD and NAFLD subjects and the majority of these two groups of individuals with  \ac{CKD}. This study aimed to compare CKD prevalence in MAFLD or NAFLD subjects and the association between the presence and severity of MAFLD and CKD and abnormal albuminuria. The authors evidenced that the severity of renal dysfunction and the prevalence of CKD stage $1$ increased progressively with the severity of MAFLD. The results of this study suggest that MAFLD better identifies subjects with CKD than NAFLD and that both MAFLD with elevated liver fibrosis scores are strongly and independently associated with CKD and abnormal albuminuria.
Compared to our work, the authors do not consider mortality to train an \ac{AI} algorithm and do not perform any explainability analysis to support physician activities.

Su Lin \emph{et al.}~\cite{lin2020comparison} proposed a comparison of MAFLD and NAFLD diagnostic criteria to validate the diagnostic criteria of MAFLD. This study aims to compare the characteristics of MAFLD and \ac{NAFLD}. 
They compared clinical parameters of MAFLD, NAFLD, and non-MD-NAFLD subjects, highlighting statistically significant differences. The main findings are that MAFLD criteria can distinguish more at-risk subjects. The MAFLD population had higher liver enzymes and more glucose and lipid metabolism disorders than NAFLD.
Compared to our study, the authors focused only on validating the diagnostic criteria of MAFLD. They did not investigate additional aspects of the disease and explainability, such as those addressed in our study.

Decraecker \emph{et al.}~\cite{decraecker2022long} suggested an evaluation of non-invasive methods in MAFLD subjects. This study aims to develop non-invasive prognostic methods to predict mortality. The authors developed several prognostic models for survival and outcome and compared the prognostic accuracy of all methods considered. The study showed that non-invasive assessment of liver fibrosis at baseline was correlated with all-cause mortality and liver-related mortality in MAFLD subjects. A predictive model (consisting of clinical parameters and measurement of liver stiffness, \ac{FIB-4} or LIVERFASt) was an excellent predictor of all-cause mortality and liver-related mortality (C-index $\approx 0.8 - 0.9$) and a good predictor of all-cause and liver-related outcomes (AUC $\approx 0.72 - 0.74$) in MAFLD.

Compared to our study, the authors focused on developing different prognostic models for survival and identifying the best predictive model for mortality in MAFLD subjects. However, the authors did not use variables readily available in healthcare databases to develop these models. 

They did not use mortality and readily available variables to train an AI algorithm and explainability analysis that can help physicians in the decision-making process in MAFLD subjects.

Nguyen \emph{et al.}~\cite{nguyen2021differential} propose a study examining the baseline characteristics and long-term outcomes of three different groups of participants with ultrasound-defined fatty liver disease. These groups are those who meet the criteria for NAFLD, but not MAFLD  (non-MAFLD NAFLD), those who meet the requirements for both NAFLD and MAFLD (NAFLD-MAFLD), and those who meet the criteria for MAFLD but not NAFLD (MAFLD non-NAFLD). The authors highlight that non-NAFLD MAFLD was independently associated with all-cause mortality compared with non-MAFLD NAFLD. In summary, persons meeting diagnostic criteria for MAFLD but not NAFLD are most likely to have advanced fibrosis. They had the highest all-cause, CVD-related, and other-cause mortality compared with those meeting the criteria for NAFLD but not MAFLD or those with both NAFLD and MAFLD. MAFLD subjects without NAFLD have more than twice the risk of mortality compared with those with NAFLD without MAFLD. 

%Compared to our study, the authors focused on comparing subjects with MAFLD and NAFLD to show that subjects with MAFLD have more than twice the mortality risk of those with NAFLD without MAFLD. 

They did not investigate the mortality to train an AI algorithm that can assist physicians in predicting mortality in MAFLD subjects. In addition, no explainability analysis is considered.

%Furthermore, the authors focused on checking mortality risk in subjects with NAFLD, MAFLD, MAFLD-NAFLD, and non-NAFLD. 
%They did not investigate mortality risk or develop a prognostic index to train an artificial intelligence algorithm to predict mortality risk in MAFLD subjects.

Kim \emph{et al.}~\cite{kim2021metabolic} developed a Cox proportional hazards model to study all-cause mortality and cause-specific mortality between MAFLD and NAFLD, with adjustments for known risk factors. The study aimed to investigate the independent longitudinal association of NAFLD and MAFLD on all-cause and cause-specific mortality in US adults.
The authors highlighted that all-cause mortality risk was higher for subjects with advanced fibrosis and MAFLD than those with advanced fibrosis and NAFLD. There was a strong association between MAFLD and all-cause mortality.
MAFLD had a $17\%$ higher risk of all-cause mortality with statistical significance, whereas NAFLD showed no association with all-cause mortality after adjustment for metabolic risk factors. These findings suggest that MAFLD is strongly associated with all-cause mortality, independently of known metabolic risk factors. 
Compared to our study, the authors focused on analyzing the independent longitudinal association of NAFLD and MAFLD on all-cause and cause-specific mortality. 

The authors do not perform any AI algorithm to train readily available variables to predict mortality and do not consider explainability analysis to assist physicians in MAFLD decision-making.

%The authors do not use mortality and readily available variables to train an artificial intelligence algorithm and explainability analysis to assist physicians in MAFLD decision-making.

Huang\emph{et al.}~\cite{huang2021nafld} investigated an evaluation of the closest association with all-cause and cause-specific mortality in MAFLD and NAFLD subjects and that drug development for MAFLD should consider ethnic differences. The authors divided the participants into four groups for survival analysis: without NAFLD or MAFLD, with NAFLD only, and MAFLD only. MAFLD increased the overall risk of all-cause mortality more than NAFLD. The risks of cardiovascular, neoplastic, and diabetes-related mortality are similar between MAFLD and NAFLD. Compared with individuals without NAFLD and MAFLD, individuals with NAFLD alone showed a reduction in total mortality and neoplasm mortality in crude. However, individuals with only MAFLD independently increased the risk of total mortality and neoplasm mortality. The risk of all-cause mortality in MAFLD was consistent among subgroups, except for race-ethnicity and whether secondary to viral hepatitis.
MAFLD showed a higher risk of all-cause mortality and equal risk of cause-specific mortality than NAFLD.

Compared to our study, the authors focused on investigating the association between all-cause and cause-specific mortality in MAFLD and NAFLD subjects. They also suggest that development for MAFLD should consider ethnic differences. 
They did not investigate mortality using variables readily available in the health dataset to train AI algorithms, and they did not perform any explainability analysis to support physicians.\\

%\textcolor{blue}{To sum up, the discussion above confirms that despite there are several contributions to the state-of-the-art, none of them analyze and evaluate the proposed techniques for...}

To sum up, the discussion above confirms that despite several contributions to state-of-the-art, none of them analyze and evaluate the proposed techniques for predicting mortality in MAFLD subjects and driving the physician in the decision-making process.
The present work represents an advancement of state of the art by introducing a novel approach to predict mortality in MAFLD subjects and using explainability techniques.

%--------------------------------------------------------------------------------------------------------------------------------------------------------
\section{The proposed approach}
\label{sec:proposed_approach}

This section describes the framework, the data used, and the proposed method to perform and explain the mortality prediction in MAFLD subjects.

\subsection{Developing a Framework for mortality prediction in MAFLD subjects}
\label{sec:framework}

In this work, a framework is proposed to classify MAFLD subjects. Then explainability techniques are applied to justify the algorithmic reasoning.
The following considerations drive the choice of experimental set in contributions: (i) to intervene early with treatment, physicians need to identify which subjects are at high mortality risk, %risk of dying, 
and (ii) physicians must understand how the algorithm predicts the target variable.

\begin{figure}[ht]
  \centering 
  \includegraphics[width=1\textwidth]{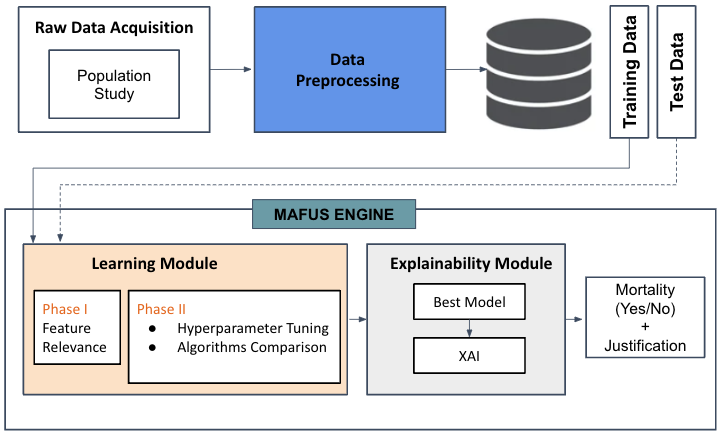}
    \caption{MAFUS classification framework.}
    \label{fig:framework}
\end{figure}

The framework is depicted in Figure~\ref{fig:framework}. The data preparation consists of the following steps: (i) acquisition of data from a population study (Section~\ref{sec:pop}) and, (ii) preprocessing operation concerning data cleaning and standardization. %As described in preprocessing Section~\ref{sec:preproc}, we removed the missing value.

To classify the data, the framework works as follows. 
%First, the inputs and the outputs (detailed in Section~\ref{sec:dataset}) are the 
For each sample, there are $24$ features and the mortality status.
%for Mortality classification.
An attribute selection tool is used to select relevant and informative attributes.
%for the classifier to guide it to the desired outcomes if all attributes are not beneficial to the classification of the dataset.

Following the data %preparation
preprocessing, a ready-to-use dataset is obtained.
Subsequently, the dataset is divided into training and test sets using the standard 80/20 method.

Our framework is developed around the MAFUS engine composed of two macro-components: (i) the \textit{Learning Module} that learns and takes care of the prediction part through the use of ML techniques and  (ii) \textit{Explainability Module} that interprets the previous model results to justify and better explain the final result.
%Given the architecture's overall figure, the learning module's outcome is the best possible prediction model trained on the dataset. 
A hyperparameter tuning is performed using a grid search technique. The output of the explainability module is the result of the
best model classification, with the addition of an explanation based on the post-hoc approach.

According to Algorithm ~\ref{alg:CMgeneration}, missing values are removed before starting the training step. A complete subset of the data is obtained and then split into train and test sets. An iterative process follows, in which the algorithm is trained, and then predictions are made. A description of the %reasoning 
operations performed by the algorithm is also provided. In the second part of the pseudocode we have two conditions: (i) if the predicted outcome is equal to $0$ (\textit{Mortality (No)}), the sample is added to set $\mathcal{A}$;
%it provides reasoning performed in predicting \textit{Mortality (No)} subject while, 
(ii) if it is equal to $1$ (\textit{Mortality (Yes)}), the sample is added to set $\mathcal{B}$.
In both cases, the SHAP-based explanation is computed.

By the end of the process, the framework is able to predict whether the patient will be \textit{Mortality (no)} or \textit{Mortality (Yes)}. In addition, \proto~ highlights the algorithmic reasoning and the contribution of each feature to the prediction. Pseudocode~\ref{alg:CMgeneration} summarizes the different steps of \proto~ framework.

\RestyleAlgo{ruled}
\SetKwInput{KwInput}{Input}

\SetKwFor{ForTrain}{for}{do}{endfor}
\SetKwFor{ForDice}{for}{do}{endfor}

\begin{algorithm}[htbp!]
    \caption{Algorithm for model training and SHAP evaluation of \proto.,}
    \label{alg:CMgeneration}

\small
\SetAlgoLined
  \textbf{Function:}
    \begin{itemize}
        \item $Clean\text{ }$ removes the missing values contained in $D_{train}$ and  $D_{test}$;
        \item $\phi_i(f,x)$ shapley values related for each feature;
    \end{itemize}
    
    \KwInput{
        \begin{itemize}
        \setlength\itemsep{0.0em}
        \item $N$ number of dataset observations; 
        \item $N_{train}$ number of observation in training set
        \item train and test datasets $D_{train}$ and $
                  D_{test}$, where $D_{train} = \{\mathcal{X}_{train}, \mathcal{Y}_{train}\}$, and $D_{test} =  \{\mathcal{X}_{test}, \mathcal{Y}_{test}\}$;
            \item target label Classifier = $f(\cdot)$;
            \item SHAP value generator $\phi(f,x_i)$ = $<$$S_1,$...$,S_z$$>$, where $z$ is the number of features;
          %  \item number of train epochs $Epochs$.
        \end{itemize}
    }
    \KwResult{
        \begin{itemize}
        \setlength\itemsep{0.0em}
        \item $\hat{y}$ represents the prediction;
        %\item $y^{*}$ represents the shapley values prediction's during the test;
        \item set $\mathcal{A}$ composed of tuples of subjects related to the samples associated with a \textit{\textit{Mortality (No)}} (i.e., $f(\mathbf{x}) = 0$);
        \item set $\mathcal{B}$ composed of tuples of subjects related to the samples associated with a \textit{\textit{Mortality (Yes)}} (i.e., $f(\mathbf{x}) = 1$);
        \item $Shapley_{values}$ = $\bigcup_{i=1}^{D_{test}} \phi(f,x_i)$;           
        %\item shap values = U i che va da 1 fino a cardinalità di Dtest di phi f x_i $<$$S_1$...$S_z$$>$
        \end{itemize}
    }
    $Preprocessing$
    \BlankLine
    
    \ForTrain{$n \gets 1$ \textbf{to} $N$}{
        $Clean\text{ }D_{train}$ and $D_{test}$\;
    }
    \BlankLine
    
    %\ForTrain{$epoch \gets 1$ \textbf{to} $Epochs$}{
        $\mathcal{X}_{train}, \mathcal{Y}_{train} \gets D_{train}$\;
    \BlankLine
    \ForTrain{$n \gets 1$ \textbf{to} $N_{train}$}{
        $\mathcal{\hat{Y}}_{train}\gets f(\mathcal{X}_{train})$\;}
    %}
    \BlankLine
    
    \ForDice{$d^{(i)} \in D_{test}$}{
    \BlankLine
    $\mathbf{x}^{(i)},y^{(i)} \gets d^{(i)}$\;
    \BlankLine
    $\hat{y}^{(i)} \gets f(\mathbf{x}^{(i)})$\;
    \BlankLine
    $Shapley_{values}\gets Shapley_{values} \cup \phi(f,x_i)$\;}
    \BlankLine
    
    \If{$\hat{y}^{(i)}=0$}{$
    \mathcal{A} \cup \{\langle\mathbf{x}^{(i)},\phi_i(f,x),\hat{y}^{(i)}\rangle\}$ \;
    }
    %\If{{$\hat{y}^{(i)}=0$}}{
    \Else{
    $\mathcal{B} \cup \{\langle\mathbf{x}^{(i)},\phi_i(f,x),\hat{y}^{(i)}\rangle\}$\;
    }
    %}
\end{algorithm}

%\textcolor{blue}{The problem we face in this work is to...}\\
%\textcolor{red}{The problem we face in this work is using machine learning techniques to predict death in MAFLD subjects. Then explainability techniques were applied to understand the algorithmic reasoning.
%The following considerations guided the choice of experimental settings reported in this paper:
%\begin{itemize}
%    \item In order to intervene early with treatment, physicians need to identify which subjects are predicted to die.
%    \item Physicians need to understand how the algorithm predicts outcomes.
%\end{itemize}
%}
%------------------------------------------------------------------------
\subsection{Population study and clinical data collection}
\label{sec:pop}

The population cohort consisted of $1,674$ subjects aged $>30$ years ($543$ women and $1131$ men) affected by MAFLD at recruitment from participants in the second follow-up of the MICOL cohort and the
NUTRIHEP cohort~\cite{mirizzi2021modified}, recruited from January 2005 and followed until December
$31$, $2020$. The baseline characteristics of the subjects are detailed in supplementary
Table~\ref{tab:baseline_tab}.

The studies have been performed at the National Institute of Gastroenterology, and the "S. De Bellis" Research Hospital in Castellana Grotte (Italy), and each participant signed an informed consent form.
All procedures are performed according to the ethical standards of the institutional research committee (MICOL study: DDG-CE-$589$/$2004$, DDG-CE-$782$/$2013$; NUTRIHEP study, DDG-CE-$502$/$2005$, and DDG-CE-$792$/$2014$) and with the $1964$ Helsinki declaration.

Participants are interviewed to collect information on sociodemographic characteristics, health status, personal history, tobacco use, dietary intake, education level, employment, and marital status.
Weight is measured with an electronic scale, SECA mechanical balance\footnote{https://us.secashop.com/products/measuring-stations-and-column-scales/seca-700/}, and height with a wall-mounted stadiometer, SECA\footnote{https://uk.secashop.com/products/height-measuring-instruments/seca-206/}. Blood pressure and BMI are measured according to international guidelines ~\cite{sever2006new}. The central hospital’s laboratory collected and processed fasting venous blood according to standard procedures~\cite{world2010guidelines}. All subjects underwent standardized ultrasonography performed by two radiologists using a Hitachi H21 Vision machine (Hitachi Medical Corporation, Tokyo, Japan) and a 3.5 MHz transducer. Steatosis is dichotomously classified as absent or present.

%------------------------------------------------------------------------
\subsection{The dataset}
\label{sec:dataset}

%The dataset consisted of $1,674$ observations. Each of them represented one MAFLD subject. 
Subjects under study had $25$ features containing continuous or categorical values of biochemical, anthropometric, and sociodemographic variables. Specifically, the $25$ columns contained the values of: 
\ac{GOT}, Weight, Hypertension, Blood lipids, \ac{SBP}, \ac{DBP}, \ac{TC}, Triglycerides, Blood Glucose, Alkaline Phosphatase, \ac{HDL-C}, \ac{LDL-C}, \ac{GPT}, \ac{GGT}, Age, \ac{HOMA}, Residual Cholesterol, \ac{BMI} as continuous variables, and Status (\textit{Mortality (No)}, \textit{Mortality (Yes)}), Education, Job, Marital Status, Diabetes condition, Smoke, Gender as categorical variables.
The dataset is represented by tabular data composed of $1,674$ observations and $25$ features, including the target variable.
The \textit{Supplementary} Table \textcolor{blue}{1} in \textit{Appendix} described the characteristics of the population under study. 

%------------------------------------------------------------------------
\subsection{Data pre-processing and features relevance analysis}
\label{sec:preproc}

%\textcolor{red}{sfruttando un framework noto nel mondo python, pandas, abbiamo potuto analizzare il dataset e rimosso tutti i valori che erano nulli.}\\
%\textcolor{red}{Using the function \textit{dropna} in pandas library, missing values (NaN) were removed randomly. Imputation of missing values was excluded because it reduces the variance of the imputed variables. Imputation shrinks standard errors, invalidating most hypothesis tests and calculating a confidence interval. Mean imputation does not preserve relationships between variables such as correlations.}
%Using the dropna function in pandas library, missing values (NaN) are removed randomly.
We parsed the dataset using a well-known Python framework, pandas\footnote{https://pandas.pydata.org/} and removed all missing values.
%Due to the reduction in the variance of imputed variables, missing values are excluded from imputed variables. The imputation process reduces standard errors, thus invalidating most hypothesis tests and calculating confidence intervals. Mean imputation does not preserve relationships between variables such as correlations.

After the missing-values removal, the number of subjects is $1,561$.
These observations are standardized according to the StandardScaler\footnote{https://scikit-learn.org/stable/modules/generated/\\sklearn.preprocessing.StandardScaler.html} technique, known in the ML literature~\cite{gadekallu2020deep} for reducing variance within datasets. Therefore, they are used to perform a feature selection analysis exploiting \ac{XGBoost} classifier.

%These $1561$ observations are standardized with the \textit{StandardScaler}\footnote{https://scikit-learn.org/stable/modules/generated/\\sklearn.preprocessing.StandardScaler.html} function (function present in scikit-learn) and used to perform a feature selection analysis exploiting \ac{XGBoost} classifier for feature importance.

%\textcolor{red}{After adopting the XGBoost classifier for feature importance, we selected the features that contributed the most to the prediction by evaluating the F1 Score value they provided to the considered outcome.}
As a result of adopting the XGBoost classifier for feature selection, we evaluated each feature's F1 score value to determine which contributed most to the prediction.
%Using XGBoost classifier for feature importance
After this selection, the features described in the previous Section are considered. The target variable considered is Status, reported as a binary variable ($0$ = \textit{Mortality (No)}, $1$ = \textit{Mortality (Yes)}).
All biochemical markers are introduced into the models as continuous variables to better reflect their natural scale. 
%The assumption behind this choice is that the effect of categorization is the loss of information.

The following nine features are identified according to their contribution, and we selected those with F1 score $>$105: Age, \ac{HDL-C}, \ac{HOMA}, \ac{BMI}, Weight, \ac{LDL-C}, Blood Glucose, Total Cholesterol, Triglycerides. Furthermore, Gender feature is added for biological importance.
Let us consider that, even if gender gets an F1 score $<$105, we added it to the feature set since in the medical domain it is considered a characteristic variable particular in liver disease~\cite{lonardo2019sex, hashimoto2011prevalence}.
%as the biochemical and anthropometric values related to the outcome differ from gender.

\begin{figure}[!h]
  \centering 
  \includegraphics[width=1\textwidth]{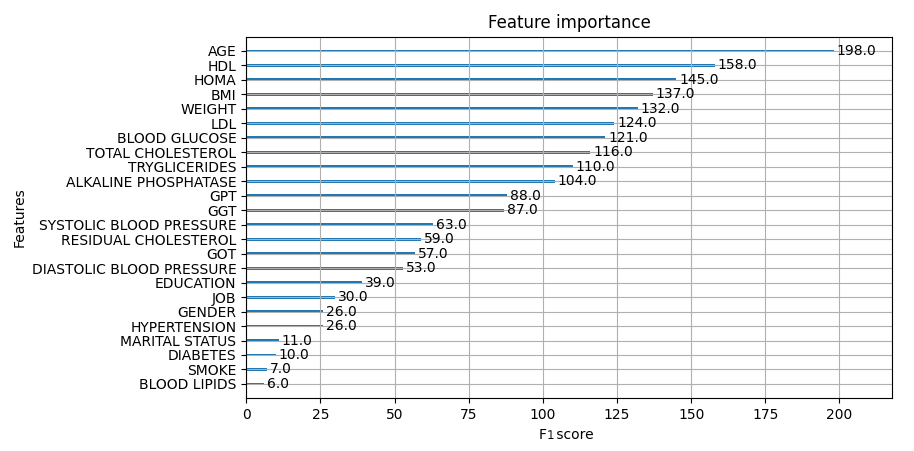}
    \caption{Contribution of each feature measured in terms of F1 score.}
    \label{fig:featrel}
\end{figure}

Figure~\ref{fig:featrel} shows the ranking of the features according to their relevance. F1 score is used to describe how each feature contributes to predicting the target variable.

%------------------------------------------------------------------------
\subsection{Classification models}
\label{sec:algo_compa}

In order to determine the best classifier to predict the mortality in MAFLD subjects, we used the following models:
\ac{MLP}~\cite{baxt1995application}; \ac{RF}~\cite{pal2005random}; \ac{SVM}~\cite{noble2006support}; Extreme Gradient Boosting (XGBoost)~\cite{chen2015xgboost}; Light Gradient Boosting Machine (LGBM)~\cite{fan2019light};

The models are developed with Python, using Scikit-learn v$1.1.2$ library\footnote{http://scikit-learn.org}.%~\cite{pedregosa2011scikit} 
The following metrics are detailed in the next Section: percentage accuracy, \ac{ROC}, precision, recall, and F1 score. 
All these five algorithms are compared to determine the best suited to predict mortality in MAFLD subjects (Section~\ref{sec:exp_res}).

%------------------------------------------------------------------------------------------
\subsection{Explainability techniques}
\label{sec:explainability}

%In order to explain the prediction, 
A post-hoc XAI approach~\cite{dovsilovic2018explainable} is used to explain how the features contribute to predicting the target variable. When prediction and explanation models are combined, there is an unavoidable trade-off between accuracy and simplicity of interpretation. This difficulty is solved by the notion of XAI.
%In XAI models, prediction and interpretation models are learned independently, such as \ac{LIME}~\cite{ribeiro2016should} and \ac{SHAP}~\cite{lundberg2017unified}. 
 XAI tools like \ac{LIME}~\cite{ribeiro2016should} and \ac{SHAP}~\cite{lundberg2017unified} are able to explain independently from the predictive model.
%These models increase interpretability without reducing prediction accuracy. 
This study uses SHAP to interpret the feature contribution to the prediction. 
As reported by Lombardi \emph{et al.}~\cite{lombardi2022robust}, we suppose that $D$ is a dataset of samples $D=\left[\left(x_1, y_1\right),\left(x_2, y_2\right), \ldots,\left(x_z, y_z\right)\right]$, where $x_i$ represents the feature vector for sample $i$ and $y_i$ represents the label for sample $i$. Assume that $f$ is a classifier, and $f_{x_i}$ is the prediction for the test instance $i$, which corresponds to the predicted label.
The goal is to explain the contribution of each feature $j$ among the $S$ features as the average marginal contribution of the feature value across all possible coalitions, i.e., , all possible sets of feature values with and without the feature $j$. Specifically, a coalition, $F$, is defined as a subset of $S, F \subseteq S$. 
Assuming $f_{x_i}(F)$ is the prediction for $f_{x_i}$ given $F$, the following equation represents the marginal contribution from adding $j$-th feature value to $F$:

\begin{equation}
    \left[f_{x_i}(F \cup j)-f_{x_i}(F)\right]
\end{equation}

To compute the exact Shapley value, all possible subsets of feature values excluding the $j$-th feature value $F \subseteq S-\{j\}$ have to be considered:

\begin{equation}
    \phi(f, x)= \sum_{F \subseteq S-\{j\}} \frac{\abs{F} !(\abs{S}-\abs{F}-1) !}{\abs{S} !}\left[f_{x_i}(F \cup j)-f_{x_i}(F)\right]
\end{equation}

where $\abs{F}!$  represents the number of permutations of feature values positioned before the $\mathrm{j}$-th feature, $(\abs{S}-\abs{F}-1)!$ represents the number of permutations of feature values that appear after the $\mathrm{j}$-th feature value, and $\abs{S}!$ is the total number of permutations~\cite{lundberg2017unified}.
The value of $\phi(f, x)$, is called \textit{SHAP value of a single feature} and, is equivalent to the Shapley value in game theory \cite{lundberg2017unified}.
The Shapley value is a value that indicates each player's participation in a cooperative game with numerous players. SHAP calculates the Shapley value of each feature as a player in the learned model. SHAP values are computed for all feature combinations, which requires exponential time.
The results of post-hoc explanation analysis applied in this paper are shown in Section~\ref{sec:XAIres}.

%In details,let $D$ be a dataset of samples, $D=\left[\left(\mathbf{x}_1, y_1\right),\left(\mathbf{x}_2, y_2\right), \ldots,\left(\mathbf{x}_{\mathbf{N}}, y_N\right)\right]$, where $\mathbf{x}_{\mathbf{i}}$ represents the feature vector for the sample $i$ and $y_i$ the corresponding label. Let $f$ be a classifier and $f_{x_i}$ the prediction for the test instance $i$, which corresponds to the predicted label.

%The goal is to explain the contribution of each feature $j$ among the $S$ features as the average marginal contribution of the feature value across all possible coalitions, i.e., all possible sets of feature values with and without the feature $j$. In particular, a coalition, $F$, is defined as a subset of $S(F \subseteq S)$. If we denote with $f_{x_i}(F)$ the prediction for $f_{x_i}$ given the subset $F$, the following equation represents the marginal contribution of adding the $j$-th feature value to $F$ :

%--------------------------------------------------------------------------------------------------------------------------------------------------------
%

\section{Experimental results}
\label{sec:exp_res}

%In this section, we present and discuss the results of predicting mortality in MAFLD subjects through our model. 
This section presents and discusses our model's experimental evaluation results for mortality prediction in MAFLD subjects.
%In addition, in Section~\ref{sec:metrics} the metrics used to evaluate machine learning models are introduced.

\textbf{Dataset Splitting.}
The dataset is split into train and test sets by the standard $80/20$ method. For reproducibility purposes, we used Scikit-learn implementation to split with a random seed set to 1\footnote{https://scikit- learn.org/stable/modules/generated/sklearn.model$\_$selection.train$\_$test$\_$split.html}.

\textbf{Decision-Maker Hyperparameter Tuning and optimization.}\label{sec:ht}
The adopted classifiers, i.e. MLP, RF, SVM, XGBoost and, LGBM described in Section~\ref{sec:algo_compa}, are tuned using a grid search exploration strategy\footnote{https://scikit- learn.org/stable/modules/generated/sklearn.model$\_$selection.GridSearchCV.html} with a $5$-fold cross-validation strategy. Due to the imbalanced nature of the datasets concerning the sensitive classes, the models optimizing the F1 score are chosen (Equation~\ref{eq:F1}). %Explored hyperparameter values are shown in Table. 
The list of explored hyperparameter values, for reproducibility, is reported in Table~\ref{tab:hyperparam_tuning}.

\begin{table}[!ht]
\caption{{Hyperparameter list, values, and type for the classification models reported in this work.}}\label{tab:hyperparam_tuning}
\centering
    \begin{tabular}{ll|cc}
    \hline
    \textbf{Algorithm}        & \textbf{Hyperparameter} & \textbf{Values}                                & \textbf{Type}        \\ \hline
    \textbf{Multilayer Perceptron}     & seed                    & \{1\}                                         & Integer              \\
                              & hidden\_layer\_sizes    & \{(sp\_randint.rvs(100, 600, 1),               & Intger                \\
                              &                         & sp\_randint.rvs(100, 600, 1),                  & \multicolumn{1}{l}{} \\
                              &                         & sp\_randint.rvs(100, 600, 1))\}                &                      \\
                              & activation              & \{tanh, relu, lbfgs\}                          & String               \\
                              & solver                  & \{sgd, adam, lbfgs\}                           & String               \\
                              & alpha                   & \{0.0001, 0.001, 0.01, 0.1, 0.9\}              & Float                \\
                              & learning\_rate          & \{constant, adaptive\}                         & String               \\ \hline
    \textbf{Random Forest}             & seed                    & \{1\}                                         & Integer              \\
                              & n\_estimators           & \{100, 200, 300, 400, 500\}                    & String               \\
                              & max\_features           & \{auto, sqrt, log2\}                           & String               \\
                              & max\_depth              & \{80, 90, 100, 110, 120, 130, 140, 150, None\} & String               \\
                              & criterion               & \{gini, entropy\}                              & String               \\
                              & class\_weight           & \{balanced\}                                   & String               \\ \hline
    \textbf{Support Vector Machines}   & seed                    & \{1\}                                         & Integer              \\
                              & class\_weight           & \{balanced\}                                   & String               \\
                              & kernel                  & \{rbf, linear\}                                & String               \\
                              & gamma                   & \{1, 0.1, 0.001, 0.0001\}                      & Float                \\ \hline
    \textbf{eXtreme Gradient Boosting} & seed                    & \{1\}                                         & Integer              \\
                              & gamma                   & \{1, 0.1, 0.01, 0.001, 0.0001\}                & Float                \\
                              & learning\_rate          & \{0.0001, 0.001, 0.01, 0.1, 1\}                & Float                \\
                              & max\_depth              & \{3, 21, 3\}                                   & Integer              \\
                              & colsample\_bytree       & \{1/10.0 for i in range(3,10)\}                & Float                \\
                              & reg\_alpha              & \{1e-5, 1e-2, 0.1, 1, 10, 100\}                & Float                \\ \hline
    \textbf{Light Gradient Boosting}   & seed                    & \{1\}                                         & Integer              \\
                              & learning\_rate          & \{0.1, 0.05\}                                  & Float                \\
                              & num\_leaves             & \{3, 10, 31, 50, 100, 200\}                    & Integer              \\
                              & reg\_alpha              & \{None, 0.01, 0.05, 0.1\}                      & Float                \\
                              & colsample\_bytree       & \{0.6, 0.8, 1\}                                & Float                \\
                              & max\_depth              & \{-1, 3, 5, 8, 10\}                            & Integer              \\
                              & reg\_lambda             & \{None, 0.01, 0.02, 0.03\}                     & Float                \\
                              & n\_estimators           & \{50, 100, 300\}                               & Integer              \\ \hline
    \end{tabular}
\end{table}

\begin{figure}[h]
\centering
    \begin{subfigure}{0.44\textwidth}
            \includegraphics[width=\linewidth]{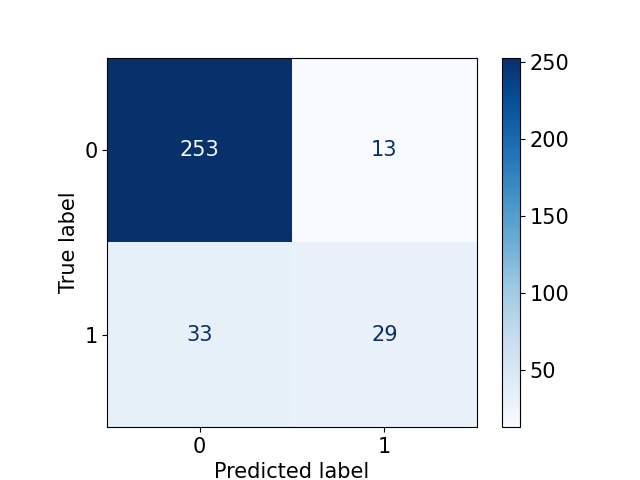}
            \caption{Confusion Matrix of MLP Classifier.}
            \label{fig:CM_MLP}
    \end{subfigure}
    \hfill
    \begin{subfigure}{0.44\textwidth}
            \includegraphics[width=\linewidth]{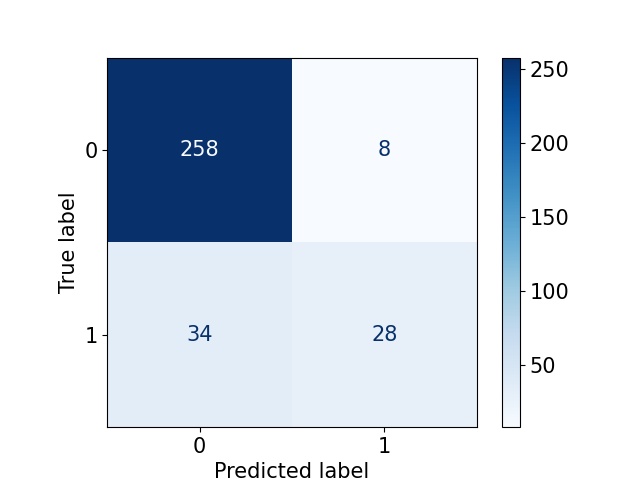}
            \caption{Confusion Matrix of RF Classifier.}
            \label{fig:CM_RF}
    \end{subfigure}
    \hfill
    \begin{subfigure}{0.44\textwidth}
            \includegraphics[width=\linewidth]{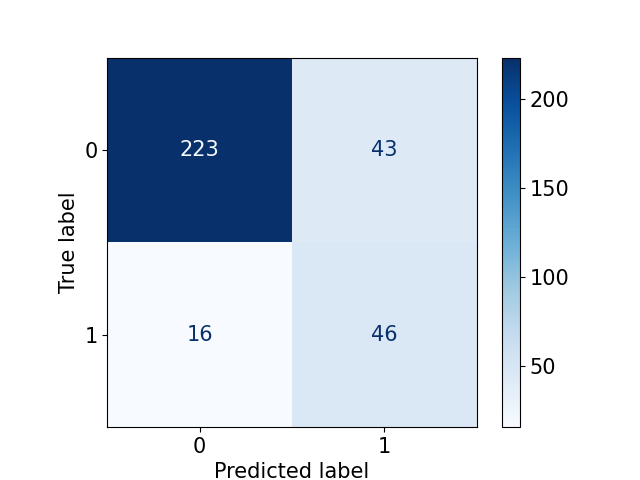}
            \caption{Confusion Matrix of SVM Classifier.}
            \label{fig:CM_SVM}
    \end{subfigure}
    \hfill
    \begin{subfigure}{0.44\textwidth}
            \includegraphics[width=\linewidth]{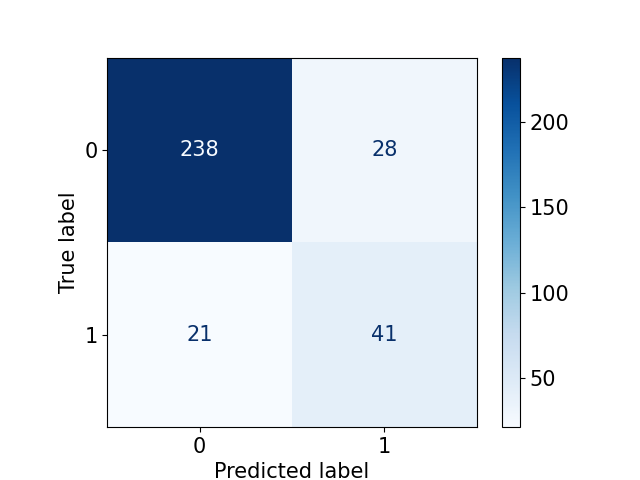}
            \caption{Confusion Matrix of XGB Classifier.}
            \label{fig:CM_XGB}
    \end{subfigure}
    \hfill
    \begin{subfigure}{0.44\textwidth}
            \includegraphics[width=\linewidth]{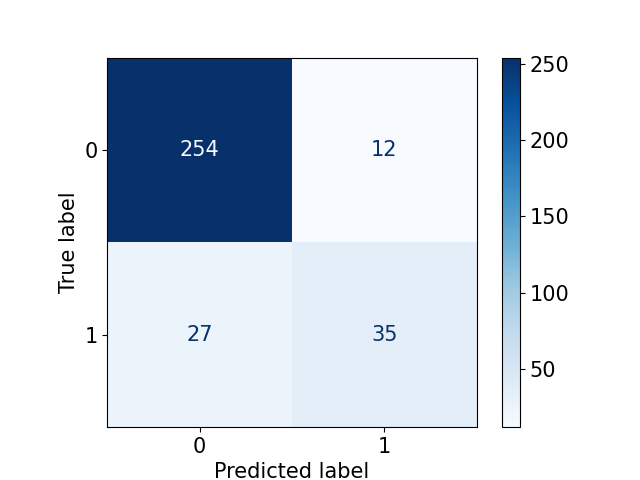}
            \caption{Confusion Matrix of LGBM Classifier.}
            \label{fig:CM_LGBM}
    \end{subfigure}
    \caption{Representation of confusion matrix values when testing LGBM, MLP, RF, SVM and XGB classifiers. $0$ = \textit{Mortality (No)}, $1$ = \textit{Mortality (Yes)}.}
\label{fig:CM_TOTALI}
\end{figure}

In order to identify the best predictive model, we compared the different algorithms in terms of the metrics reported in Section~\ref{sec:metrics}. Table~\ref{tab:results} shows the performance of each algorithm in predicting mortality.
%After the hyperparameter tuning, the metrics shown in Section~\ref{sec:metrics} are compared for each algorithm.
In addition, the \ac{CM} is evaluated in Figure~\ref{fig:CM_TOTALI}, to highlight the number of subjects misclassified during the testing phase.

\begin{table}[h]
    \caption{Results for the mortality prediction of the compared algorithms.\\ No/Yes indicate mortality condition.}\label{tab:results}
    \centering
    \begin{tabular}{lllllll}
    \hline
                               &   & \multicolumn{5}{c}{\textbf{Classifier}}                                  \\ \cline{3-7} 
                               &   & \textbf{RF} & \textbf{XGB} & \textbf{MLP} & \textbf{LGBM} & \textbf{SVM} \\ \hline
    \multirow{2}{*}{Precision} & No & 0.88        & 0.92         & 0.88         & 0.90          & \textbf{0.93}         \\
                               & Yes & \textbf{0.78}        & 0.59         & 0.69         & 0.74          & 0.52         \\
    \multirow{2}{*}{Recall}    & No & \textbf{0.97}        & 0.89         & 0.95         & 0.95          & 0.84         \\
                               & Yes & 0.45        & 0.66        & 0.47         & 0.56          & \textbf{0.74}         \\
    \multirow{2}{*}{F1 score}  & No & 0.92        & 0.91         & 0.92         & \textbf{0.93}          & 0.88         \\
                               & Yes & 0.57        & 0.63         & 0.56         & \textbf{0.64}          & 0,61         \\
    Accuracy                   &   & 0.87        & 0.85         & 0.86         & \textbf{0.88}          & 0.82         \\
    AUC                        &   & 0.88        & 0.85         & \textbf{0.90}        & 0.89          & \textbf{0.90}         \\ \hline
    \end{tabular}
\end{table}

%\clearpage

%------------------------------------------------------------------------
\subsection{Evaluation metrics}
\label{sec:metrics}

In this Section, accuracy-based metrics are introduced. They are mainly based on the CM.
%, which quantifies how many samples are correctly classified or misclassified for both the negative and positive classes. 
%For self-consistency, this section details all the considered metrics. 
%Some are just recalled, reporting the formulas.
The first metric is Accuracy, which quantifies the overall ratio of correct classifications for the whole dataset:
\begin{align}
\label{eq:acc}
    Accuracy =\frac{TP+TN}{TP+TN+FP+FN}
\end{align}
where $TP$, $TN$, $FP$, and $FN$ represent the number of true positive, true negative, false positive, and false negative predictions, respectively. 
The Recall metric measures the ratio of correct positive classifications among the total number of positive samples:
\begin{align}
\label{eq:rec}
    Recall =\frac{TP}{TP+FN}
\end{align}
The Precision measures the ratio of correct positive classifications among the total positive classifications:
\begin{align}
\label{eq:prec}
    Precision =\frac{TP}{TP+FP}
\end{align}
The F1 score is the harmonic mean between recall and accuracy:
\begin{align}
\label{eq:F1}
    F 1 =\frac{2}{\frac{1}{\text { Precision }}+\frac{1}{\text { Recall }}}=2 \cdot \frac{\text { Precision } \cdot \text { Recall }}{\text { Precision }+\text { Recall }}
\end{align}
The F1 score combines precision and recall into a single metric. This metric is useful when dealing with imbalanced data. The Area Under the Receiver Operating Characteristic Curve (AUC) is a metric that measures the capability of a classifier to separate the positive class from the negative one. It is formulated as follows:
\begin{align}\label{AUC}
    AUC=\frac{\sum_{x^{-} \in X^{-}} \sum_{x^{+} \in X^{+}}\left(\mathbbm{1}\left(f\left(x^{-}\right)<f\left(x^{+}\right)\right)\right)}{|X|^{-}+|X|^{+}} \quad \text { where } \mathbbm{1}(\cdot)=1 \text { if } f\left(x^{-}\right)<f\left(x^{+}\right) \text {else } \mathbbm{1}(\cdot)=0
\end{align}
where $X^{+}$ is the set of positive samples, $X^{-}$ is the set of negative samples, $f(\cdot)$ is the result of model prediction, and $\mathbbm{1}(\cdot)$ an indicator function~\cite{calders2007efficient}.\\

%------------------------------------------------------------------------
\subsection{Best model performance evaluation}
\label{sec:best_model}

In this experimental session, we compared the performances of the classifiers. 
As we can observe in Figure \ref{fig:CM_RF} and Table \ref{tab:results}, RF obtains the highest value of recall ($0.97$), the least number of misclassified in the prediction of class \textit{Mortality(No)}, and the best value of precision, in the prediction of \textit{Mortality(Yes)}($0.78$). However, the algorithm is not able to correctly discriminate subjects in the \textit{Mortality (Yes)} class, resulting the worst for this task. 
LGBM obtains the best value of F1 score in predicting both classes \textit{Mortality(No)}($0.93$); \textit{Mortality(Yes) ($0.64$)}, the best value of accuracy on the test set ($0.88$) and good discriminative power in predicting \textit{Mortality (Yes)}. 
MLP, on the other hand, obtains the best AUC value ($0.90$) with the SVM algorithm. However, MLP as RF cannot correctly discriminate subjects belonging to \textit{Mortality (Yes)}.
XGB, while performing well on average, does not provide any preponderant metrics compared to the other algorithms. However, it is very balanced and has good discriminative power in predicting \textit{Mortality (Yes)}. SVM instead obtains the best recall values in predicting \textit{Mortality (Yes)} ($0.74$), the best precision value in predicting \textit{Mortality (No)} ($0.93$), and is the algorithm with the best discriminative power in predicting \textit{Mortality (Yes)} as well as having the highest AUC (along with MLP). The goal was to maximize the F1 score in predicting \textit{Mortality (Yes)} class as the dataset was unbalanced, and we wanted to minimize the number of misclassified in \textit{Mortality (Yes)}.
However, as discussed earlier, although LGBM provided the highest F1 score (in the prediction \textit{Mortality (Yes)}) of $0.64$ (Table~\ref{tab:results}), we found that SVM is better in predicting \textit{Mortality (Yes)} than other algorithms, despite having a $0.61$ lower F1 value than LGBM.
Since the main goal of this work is to be accurate in identifying subjects with the highest mortality risk in order to allow the physician to intervene early with the lifestyle-change recommendations, the model with the smallest number of errors in predicting Mortality (Yes) is the best. Accordingly, SVM that shows 16 samples misclassified (cfr. Figure \ref{fig:CM_SVM}) is our best choice.

\begin{figure}[h]
  \centering 
  \includegraphics[width=0.8\textwidth]{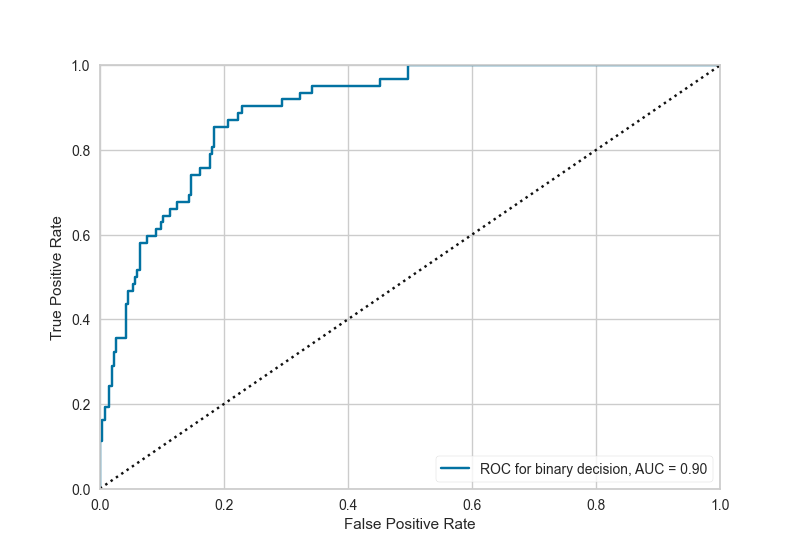}
    \caption{Representation of ROC curve and AUC SVM during the test.}
    \label{fig:AUC}
\end{figure}

%We also looked at the \ac{CM} and ROC curves to complete the analysis.
Figure~\ref{fig:AUC} shows the ROC curve with an AUC of $0.90$ obtained from SVM testing. As a measure of accuracy, the area under the ROC (AUC) is used to estimate the test's discriminant power. 

In particular, SVM has a high AUC, which indicates that  \textit{Mortality (No)} and \textit{Mortality (Yes)} are correctly classified.

%\textcolor{blue}{We used the CM to reinforce the choice of SVM as the best model by quantifying the number of samples classified correctly in \textit{Mortality (Yes)} class. Figure~\ref{fig:CM_SVM} highlights the performance of the SVM during the testing phase in predicting the considered target variable. As shown in Figure~\ref{fig:CM_SVM}, the SVM Confusion Matrix is more effective during the prediction phase than the other algorithms in detecting \textit{Mortality (Yes)}.}

%------------------------------------------------------------------------
\subsection{Explainability results}
\label{sec:XAIres}

We perform an XAI analysis on the best model results to summarize what we briefly described above. 
XAI analysis aims to analyze and compare the model's detection (or decision) strategy to increase trust and understand algorithmic reasoning and contributions of each feature to prediction.
XAI analysis is carried out using Python's \textit{SHAP}\footnote{https://shap.readthedocs.io/en/latest/index.html} library, after checking the behavior of the SVM (Section~\ref{sec:best_model}).

\begin{figure}[ht]
  \centering 
  \includegraphics[width=0.8\textwidth]{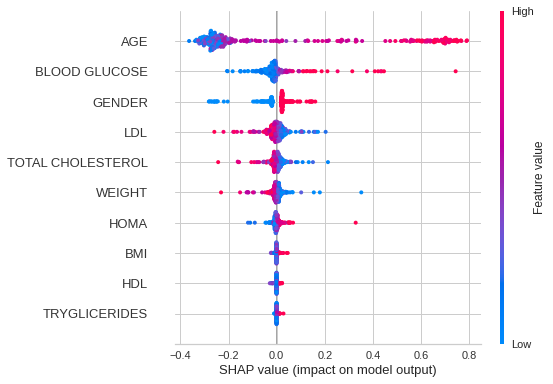}
    \caption{Representation of shaply values of anthropometric and biochemical parameters considered.}
    \label{fig:scatter}
\end{figure}

Figure~\ref{fig:scatter} shows the synthesis global diagram using Shapley values of anthropometric and biochemical parameters. There is a relationship between the importance of features and their effects.

The position on the $y$-axis is determined by the feature while on the $x$-axis by the Shapley value, as detailed in Section~\ref{sec:explainability}. From low to high, from blue to red respectively, the color represents the value of the feature.
By jittering overlapping points along the y-axis, we obtained a sense of the distribution of Shapley values across features.  
%The features are ranked according to their importance.\\
In particular, the synthesis global diagram illustrates the following:

\begin{itemize}
    \item \textit{Feature Selection}: variables ranked in descending order of importance;
    \item \textit{Impact}: horizontal position shows whether the effect of that value is associated with a higher or lower prediction;
    \item \textit{Value}: color shows whether that variable is high or low for that observation. The red color means high values and the blue low values. The change in color of the dot shows the value of the feature;
    \item \textit{Correlation}: shows the relationship of each characteristic with the target variable.
\end{itemize}

We assume the direction of Shapley to be independent of XGboost ranking. Shapley values are required to provide the prediction direction from the clinical perspective.

As shown in Figure~\ref{fig:scatter}, in our experiments the $Age$ feature is the most important, while the $Triglyceride$ feature is the least important.

After identifying $Age$ as the most significant feature in the prediction, a study of the dependence of SHAP features is conducted, analyzing the dependencies between all other features and $Age$.
SHAP dependence graphs show the expected results of a method. In this way, you can show how a feature affects model output, thus demonstrating the model's dependence on it. More specifically in a SHAP dependence graph:
\begin{itemize}
    \item Each dot is a single prediction (row) from the dataset;
    \item x-axis is the value of the feature (from the $X$ matrix);
    \item y-axis is the SHAP value for that feature, which represents how much knowing that feature's value changes the output of the model for that sample's prediction.
    \item By default, the color corresponds to a second feature that may interact with the feature we are plotting. If an interaction effect is present between this other feature and the feature we are plotting it will show up as a distinct vertical pattern of coloring.
\end{itemize}

%\textcolor{red}{(Da capire se lasciare questo)To indicate the feature's marginal effect on the model output and the relationship to the feature with which it interacts most.}

\begin{figure}[!ht]
\centering
    \begin{subfigure}{0.45\textwidth}
            \includegraphics[width=\linewidth]{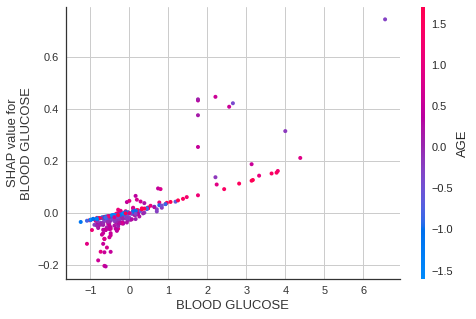}
            \caption{Dependence plot of Blood Glucose vs. Age SHAP value in MAFUS.}
            \label{fig:DEP_Bg}
    \end{subfigure}
    \hfill
    \begin{subfigure}{0.45\textwidth}
            \includegraphics[width=\linewidth]{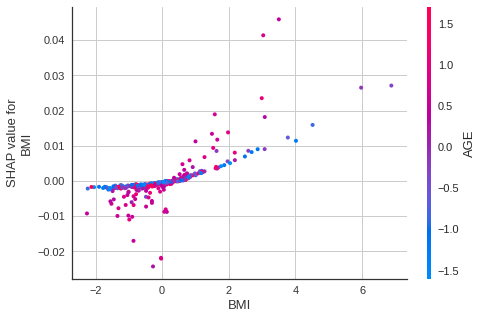}
            \caption{Dependence plot of BMI vs. Age SHAP value in in MAFUS.}
            \label{fig:DEP_BMI}
    \end{subfigure}
    \hfill
    \begin{subfigure}{0.45\textwidth}
            \includegraphics[width=\linewidth]{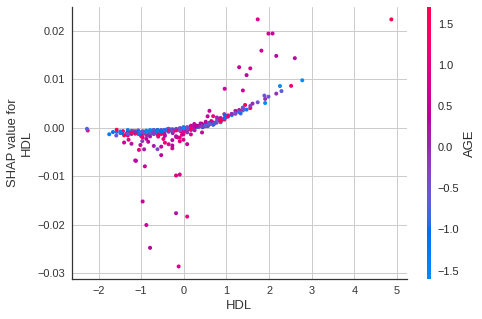}
            \caption{Dependence plot of HDL vs. Age SHAP value in in MAFUS.}
            \label{fig:DEP_HDL}
    \end{subfigure}
    \caption{Representation of dependence plot for AGE vs. BLOOD GLUCOSE, BLOOD GLUCOSE vs. AGE and, BMI and HDL vs. AGE (the feature with largest contribution in the prediction).}\label{fig:Dep_Tot_1}
\end{figure}

\begin{figure}[!ht]
\centering
    \begin{subfigure}{0.45\textwidth}
            \includegraphics[width=\linewidth]{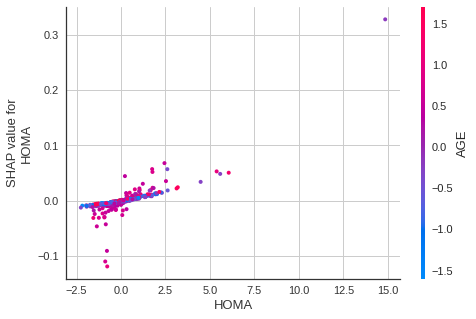}
            \caption{Dependence plot of HOMA vs. Age SHAP value in in MAFUS.}
            \label{fig:DEP_Homa}
    \end{subfigure}
    \hfill
    \begin{subfigure}{0.45\textwidth}
            \includegraphics[width=\linewidth]{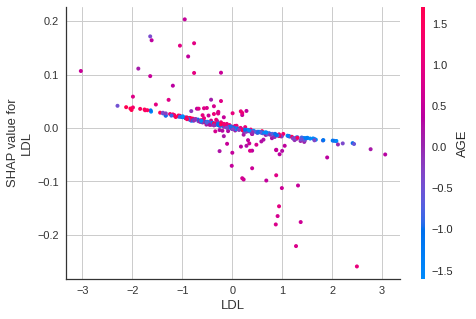}
            \caption{Dependence plot of LDL vs. Age SHAP value in in MAFUS.}
            \label{fig:DEP_LDL}
    \end{subfigure}
    \hfill
    \begin{subfigure}{0.45\textwidth}
            \includegraphics[width=\linewidth]{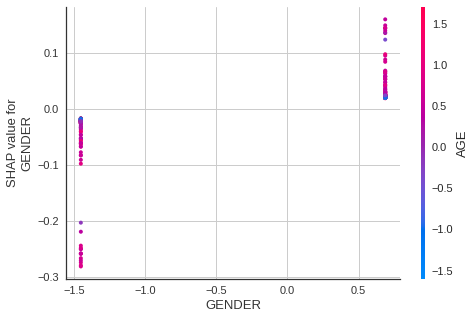}
            \caption{Dependence plot of Gender vs. Age SHAP value in MAFUS.}
            \label{fig:DEP_Sex}
    \end{subfigure}
    \hfill
    \begin{subfigure}{0.45\textwidth}
            \includegraphics[width=\linewidth]{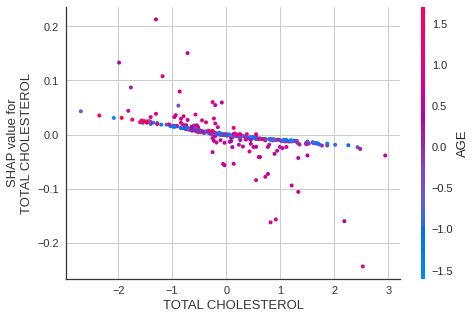}
            \caption{Dependence plot of Total Cholesterol vs. Age SHAP value in in MAFUS.}
            \label{fig:DEP_TotC}
    \end{subfigure}
    \hfill
    \begin{subfigure}{0.45\textwidth}
            \includegraphics[width=\linewidth]{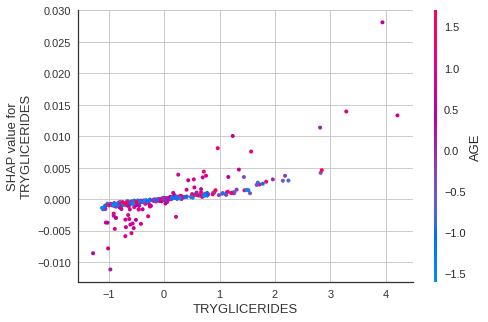}
            \caption{Dependence plot of Tryglicerides vs. Age SHAP value in MAFUS.}
            \label{fig:DEP_Try}
    \end{subfigure}
    \hfill
    \begin{subfigure}{0.45\textwidth}
            \includegraphics[width=\linewidth]{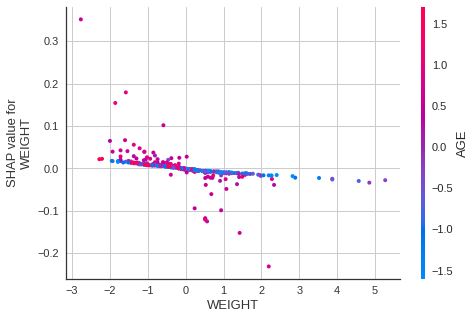}
            \caption{Dependence plot of Weight vs. Age SHAP value in MAFUS.}
            \label{fig:DEP_Weig}
    \end{subfigure}
    \caption{Representation of dependence plot for HDMA, LDL, GENDER and TOTAL CHOLESTEROL, TRYGLICERIDES and, WEIGHT vs. AGE (the feature with largest contribution in the prediction).}\label{fig:Dep_Tot_2}
\end{figure}

Figures~\ref{fig:Dep_Tot_1} and~\ref{fig:Dep_Tot_2} show the dependence of each feature.
%to feature $Age$. 
Specifically, we evaluated the most indicative features. They are illustrated below.
%Dependency graphs are presented below with a description of the different dependencies.Figure~\ref{fig:DEP_Age} shows that low values of feature $Age$ and feature $BMI$ contribute to prediction toward \textit{Mortality (No)}. As feature $Age$ and feature $BMI$ increase, the contribution in prediction shifts toward \textit{Mortality (Yes)}.In particular, Figure~\ref{fig:DEP_Age} shows that low values on the x-axis and high importance on the y-axis generally contribute to the prediction of \textit{Mortality (No)}. With increased feature x and y axes, prediction shifts to \textit{Mortality (Yes)}. Nonetheless, it is interesting that the prediction tends to shift toward \textit{Mortality (Yes)} for average features on the x and y-axes, rispectively. Figure~\ref{fig:DEP_Age} shows that generally low values of feature on $x-axis$ and high value of feature on y-axis contribute to prediction toward \textit{Mortality (No)}. As feature x-axis and feature y-axis increase, the contribution in prediction shifts toward \textit{Mortality (Yes)}.However, it is interesting to note that for average feature values on the $x-axis$ and $y-axis$ the prediction tends to be shifted toward \textit{Mortality(Yes)}. 

%Figure~\ref{fig:DEP_Sex} show that similar value of feature on x-axis given different impact on the prediction. Furthermore for male (Value on x-axis $>$ $0.5$) it can be noted that the prediction shifts toward \textit{Mortality (Yes)} as age increases.

According to Figure~\ref{fig:DEP_Sex}, similar features on the x-axis have different impacts on prediction. As the age of the male increases (Value on the x-axis $>$ $0.5$), the prediction shifts toward \textit{Mortality (Yes)}.

%Figures~\ref{fig:DEP_Bg},~\ref{fig:DEP_BMI}, ~\ref{fig:DEP_HDL}, ~\ref{fig:DEP_Homa}, ~\ref{fig:DEP_LDL}, ~\ref{fig:DEP_TotC},~\ref{fig:DEP_Try}, ~\ref{fig:DEP_Weig} shows that generally low values of feature on $x-axis$ and high value of feature on $y-axis$ contribute to prediction toward \textit{Mortality (No)}. As feature $x-axis$ and feature $y-axis$ increase, the contribution in prediction shifts toward \textit{Mortality (Yes)}. Also can be notice similar value of feature on $x-axis$ given different impact on the prediction.
In Figures~\ref{fig:DEP_Bg},~\ref{fig:DEP_BMI}, ~\ref{fig:DEP_HDL}, ~\ref{fig:DEP_Homa}, ~\ref{fig:DEP_LDL}, ~\ref{fig:DEP_TotC},~\ref{fig:DEP_Try}, ~\ref{fig:DEP_Weig} it is shown that low values of features on the x-axis and high values of features on y-axis contribute to predicting \textit{Mortality (No)}. As the x and y-axes of the feature increase, \textit{Mortality (Yes)} becomes more of a contributor to the prediction. The x-axis also shows a similar value for features with different impacts on prediction.

%Figure~\ref{fig:DEP_Bg} shows that low values of feature $Blood Glucose$ and high value of feature $Age$ contribute to prediction toward \textit{Mortality (No)}. As feature $Blood Glucose$ and feature $Age$ increase, the contribution in prediction shifts toward \textit{Mortality (Yes)}.
%Based on Figure~\ref{fig:DEP_Bg}, low values of feature $Blood Glucose$ and high values of feature $Age$ contribute to prediction of \textit{Mortality (No)}. Feature $Blood Glucose$ contributes less to prediction as feature $Age$ increases \textit{Mortality (Yes)}.

%Figure~\ref{fig:DEP_BMI} shows that low values of feature $BMI$ and high value of feature $Age$ contribute to prediction toward \textit{Mortality (No)}. As feature $BMI$ and feature $Age$ increase, the contribution in prediction shifts toward \textit{Mortality (Yes)}.

%Figure~\ref{fig:DEP_HDL} shows that low values of feature $LDL$ and high value of feature $Age$ contribute to prediction toward \textit{Mortality (No)}. As feature $LDL$ and $Age$ feature  increase, the contribution in prediction shifts toward \textit{Mortality (Yes)}.

%Figure~\ref{fig:DEP_Homa} shows that low values of feature $Homa$ and high value of feature $Age$ contribute to prediction toward \textit{Mortality (No)}. As feature $Homa$ and feature $Age$ increase, the contribution in prediction shifts toward \textit{Mortality (Yes)}.

In addition, we observe how the SVM algorithm worked by randomly selecting two subjects from the \textit{Mortality (No)} and \textit{Mortality (Yes)} classes, and obtaining a prediction that verifies $No$  and $Yes$ respectively.

\begin{figure*}[!htbp]
  \centering 
  \includegraphics[width=1\textwidth]{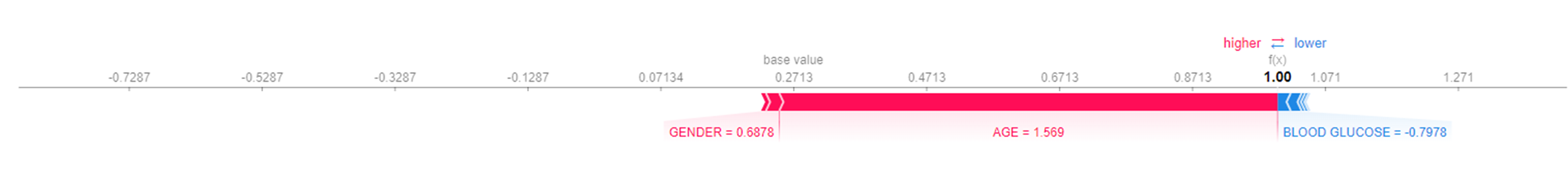}
    \caption{Application of SHAP for prediction of \textit{Mortality (Yes)} during the test.}
    \label{fig:pred_1}
\end{figure*}

Figure~\ref{fig:pred_1} displayed the positive and negative SHAP values on the left and right sides as if competing against each other.
In particular, it shows which characteristics had a greater impact on a prediction of \textit{Mortality (Yes)}.

As shown in Figure~\ref{fig:pred_1}, the characteristics $Gender$ and $Age$ are both positive predictors, whereas $Blood$ $Glucose$ is negative.

\begin{figure*}[ht]
  \centering 
  \includegraphics[width=1\textwidth]{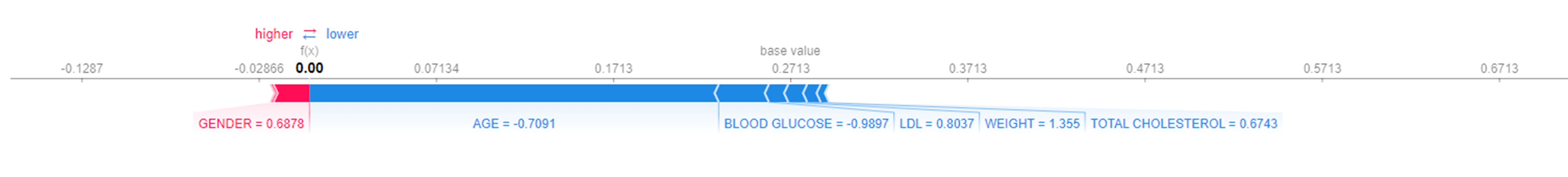}
    \caption{Application of SHAP for prediction of \textit{Mortality (No)} during the test.}
    \label{fig:pred_0}
\end{figure*}

Figure~\ref{fig:pred_0}  shows which characteristics had a greater  on a prediction of \textit{Mortality (No)}.

As shown in Figure~\ref{fig:pred_0}, the feature $Gender$, the set of features $Age$, $Blood$ $Glucose$, $LDL$, $Weight$ and, $Total$ $Cholesterol$ have a positive and negative impact on the prediction, respectively.
%--------------------------------------------------------------------------------------------------------------------------------------------------------
%\section{Discussion}\label{sec:discussion}
\section{Conclusion and further directions}
\label{sec:conclusion}

This research proposes a framework to classify mortality in MAFLD subjects.
The framework consists of two main components: an $AI Analysis$ module and a $XAI Analysis$, both based on machine learning techniques (MLTs).
As part of this work, we compared various machine learning algorithms to determine which is best at predicting mortality among MAFLD subjects.

In all cases, the algorithms under study performed well in terms of accuracy. A good F1 score and the lowest number of misclassifications for the Mortality (Yes) class are considered when choosing the best algorithm. XGboost, LGBM, and SVM had the best F1 scores with $0.62$, $0.64$, and $0.61$ respectively. 

%It was found from the results that SVM, compared with other algorithms under study, produced few prediction errors in the testing phase, although they have low accuracy and a lower F1 score in class 1 prediction than XGboost and LGBM.\\
According to the results (Section~\ref{sec:exp_res}), SVM produced the fewest prediction errors in the testing phase, despite having a lower accuracy and  F1 score than XGboost and LGBM.

%Therefore, it was essential to train an SVM to identify the mortality of MAFLD subjects and explain the prediction of the SVM algorithm so that the clinician could understand the reasoning that led the algorithm toward a given prediction. The goal was to preemptively identify subjects with the outcome "1" in order to be able to target them for early treatment in such a way modify that this condition. 

To determine the mortality risk in MAFLD subjects, an SVM is trained and explained so that the clinician can understand how the algorithm came to the prediction. Identifying subjects with target variable $1$ (\textit{Mortality (Yes)}) is the goal to allow the physician to intervene early with lifestyle-change recommendations (suggesting Diet and Physical Activity) according to Curci \emph{et al.}~\cite{curci2022effect}.
%Furthermore, the XGB and LGBM models have reasonable F1 Scores of 0.63 and 0.64 respectively, in predicting \textit{Mortality Yes}. This study aims to find an algorithm that predicts \textit{Mortality Yes} with the fewest misclassifications. As a result, SVM is selected as the best model. 

It is important to note that the dataset used for training and testing ML algorithms is unbalanced and contains a few subjects classified as \textit{Mortality (Yes)}. This unbalanced dataset does not allow the algorithm to predict this class well.
Another limitation is that no MAFLD mortality classifier has been developed in other studies, so we cannot compare our results to other baselines.

The novelty of the proposal, as well as its remarkable properties and flexibility, has the potential to pave the way for further research.

In future work, \ac{DL} algorithms will be assessed and analyzed in relation to the prediction of mortality in MAFLD subjects. In addition, other approaches based on feature relevance can be studied.
Other datasets will be studied to predict mortality in MAFLD subjects, especially local ones, where the proposed method can be applied. The number of observations in the dataset used in the study will be increased, and a DL algorithm will be developed to increase predictive performance. Subsequently, a web-based application based on ML/DL algorithm may be developed in order to provide an easy-to-use tool for the physician.
%------------------------------------------------------------------------------------------
%\section{Discussion and Further Directions}\label{sec:discussion}

%------------------------------------------------------------------------------------------
%\section{Conclusion and Future Work}\label{sec:conclusion}
%\textcolor{blue}{In this paper, we presented a...}
%------------------------------------------------------------------------------------------
\backmatter
%----------------------------------------------------------------------------------------------------------------------
%\bmhead{Supplementary infoto
%If your article has accompanying supplementary file/s please state so here. 
%Authors reporting data from electrophoretic gels and blots should supply the full unprocessed scans for key as part of their Supplementary information. This may be requested by the editorial team/s if it is missing.
%Please refer to Journal-level guidance for any specific requirements.
\bmhead{Acknowledgments}
{This work was partial support of the projects: Italian P.O. Puglia FESR 2014 -- 2020 (project code $6$ESURE$5$) `SECURE SAFE APULIA', Fincons CdP3, PASSPARTOUT, Servizi Locali 2.0, ERP4.0. We thanks to Micol and Nutriep group. We thank Dr. Alberto Rubén Osella for authorizing the use of the data. }

\noindent
%If any of the sections are not relevant to your manuscript, please include the heading and write `Not applicable' for that section. 

\bigskip

\bibliographystyle{elsarticle-num-names}
\bibliography{biblio}


\end{document}